\documentclass[11pt]{article}

\usepackage[utf8]{inputenc}
\usepackage{xcolor}

\usepackage[numbers,sort&compress]{natbib}
\usepackage[nottoc]{tocbibind}
\usepackage{latexsym,amssymb,amsmath}
\usepackage{amsthm}
\usepackage{hyperref} 

\usepackage{subcaption}

\usepackage{graphicx}

\newcommand{\lean}[1]{{\bf \color{blue} #1}}

\usepackage{authblk} 
\usepackage{footmisc} 

\title{Lagrangian neural networks for nonholonomic mechanics}

\author[*]{Viviana Alejandra Díaz}
\author[**]{Leandro Martín Salomone}
\author[**]{Marcela Zuccalli}
\affil[*]{Departamento de Matemática, Universidad Nacional del Sur, Bahía Blanca, Argentina}
\affil[**]{Centro de Matemática de La Plata, Universidad Nacional de La Plata, La Plata, Argentina}

 \DeclareGraphicsExtensions{.png,.pdf}

\begin{document}

\maketitle


\begin{abstract}

Lagrangian Neural Networks (LNNs) are a powerful tool for addressing physical systems, par\-ti\-cu\-lar\-ly those governed by conservation laws. LNNs can parametrize the Lagrangian of a system to predict trajectories with nearly conserved energy. These techniques have proven effective in un\-cons\-trai\-ned systems as well as those with holonomic constraints. In this work, we adapt LNN techniques to mechanical systems with nonholonomic constraints. We test our approach on some well-known examples with nonholonomic constraints, showing that incorporating these restrictions into the neural network's learning improves not only trajectory estimation accuracy but also ensures adherence to constraints and exhibits better energy behavior compared to the unconstrained counterpart. 
\end{abstract}


\section{Introduction} 


The laws of motion of a Lagrangian system are determined by the principle of stationary action, also known as Hamilton's principle. This principle states that the action is minimal (or stationary) throughout a mechanical process. From this statement, the differential equations known as Euler-Lagrange equations are derived. 
If the Lagrangian function of a given mechanical system is known, then Euler-Lagrange equations establish the relationship between accelerations, velocities, and positions; that is, the system dynamics are obtained from Euler-Lagrange equations. Hence, the goal of Lagrangian mechanics is to write an analytic expression for the Lagrangian function in appropriate generalized coordinates and then develop the Euler-Lagrange equations symbolically into a system of second-order differential equations whose solutions give the system's trajectory.

In many cases, even when the Euler–Lagrange equations are available, the solutions are not given in analytical or explicit form. In such cases, numerical integrators can be used to estimate the trajectories of a mechanical system. However, these methods may yield trajectories that behave poorly with respect to certain physical observables, such as energy. As an alternative, geometric integrators can be employed, as they are known to preserve energy (see, for instance, \cite{CelledoniPHD19}). Nevertheless, their accuracy may degrade over long time intervals. An even more challenging situation arises when an analytical expression for the Lagrangian is unavailable or difficult to handle, leaving us without a system of equations to solve.

In the presence of constraints, mechanical systems can be broadly classified as holonomic or nonholonomic, depending on the nature of the constraints. Holonomic constraints can be expressed as algebraic equations involving only the generalized coordinates. These constraints are integrable and reduce the dimension of the configuration space; the dynamics of such systems can be derived from a classical variational principle by incorporating the constraints through Lagrange multipliers. This framework preserves key geometric properties, such as symplecticity and, under suitable conditions, energy conservation.

In contrast, nonholonomic systems are characterized by non-integrable, velocity-dependent constraints, and differ fundamentally from holonomic systems both in their geometric structure and in the formulation of their equations of motion. As a result, the study of nonholonomic mechanical systems is a classical subject that has remained highly active over the past few decades, driven by both practical applications—such as robotics, locomotion, and control—and by questions of intrinsic mathematical interest. Representative works in the field include \cite{carira93,CMR01b,dLdD96b,EKMR05,neimarkfufaev,marle98,bloch,batesGmD96,batessniatycki,BKMM96,Mestdag,cushmanKSB,CdDdLM00}.

Unlike holonomic constraints, which restrict the configuration space, nonholonomic constraints define particular admissible velocities. As a consequence, the dynamics of nonholonomic systems are not derived from Hamilton’s principle but rather from the Lagrange-d'Alembert principle (when the constraint is linear in velocities) or the Chetaev principle (for generic constraints, not necessarily linear in velocities), and they do not, in general, preserve symplectic structure or energy. This intrinsic difference poses serious challenges for standard numerical integrators, which often fail to preserve the constraints or yield long-term energy drift. In fact, many classical integration schemes do not respect the geometry of nonholonomic flows, leading to qualitative inaccuracies in the simulation of such systems.

In recent years, there has been an increasing interest in using neural networks to address different issues of mechanical systems (see for example \cite{Simplifying},\cite{NEURIPS2019_26cd8eca},\cite{lutter2019deep},\cite{Hamiltonian},\cite{Toth2019HamiltonianGN}). In this line, Lagrangian Neural Networks were introduced in \cite{Cranmer2020LagrangianNN} as an enhancement over other types of neural networks used in mechanical systems that do not preserve physical laws, providing a tool for scenarios where, for example, equations of motion are not available to get the actual trajectory. This method assumes that the Lagrangian of a mechanical system, a scalar function, can be parametrized using a neural network and be learned directly from the system's data. That is, the goal of LNNs is to predict the Lagrangian function of a system based on data about its positions and velocities. 
These models incorporate physical structure into the learning process and have shown improved performance over standard integrators in preserving key properties such as energy. 
This approach aims to solve the system's dynamics from a Lagrangian learned by a neural network while ensuring the preservation of some specific physical properties. Given these advantages, it is natural to explore whether a properly adapted version of LNNs can offer similar benefits in the nonholonomic setting, where traditional methods struggle.

From a theoretical standpoint, it is well known that the same laws do not always govern the dynamics of constrained systems as those of unconstrained systems. Since LNNs, as introduced in \cite{Cranmer2020LagrangianNN}, are trained using the equations of motion for unconstrained systems, it is natural to ask whether such models can still learn meaningful Lagrangian functions when applied to systems with constraints. Some previous works, such as \cite{Simplifying} and \cite{ZhongBenchmarking}, have addressed holonomic constraints.

The motivation for this work is to investigate whether a targeted adaptation of LNNs can more effectively handle nonholonomic constraints. As shown in the examples we present, standard LNNs fail to preserve the constraint along the trajectories generated by the model, even when starting from initial states that satisfy the constraint.

To address this issue, we propose modifying the loss function of the Lagrangian neural network to explicitly account for nonholonomic constraints during training, when such constraints are present in the mechanical system. In the examples studied, this modification yields in solutions that more closely adhere to the system’s physical properties -particularly in terms of energy conservation for most of the examples considered, and most notably in the preservation of the nonholonomic constraint. By doing so, we aim to extend the applicability of Lagrangian-based learning to this important and challenging class of mechanical systems.

\section{Lagrangian mechanics} 

A Lagrangian mechanical system is defined as a pair \((Q, L)\), where \(Q\) is an \(n\)-dimensional differentiable manifold, known as the configuration space, and \(L: TQ \rightarrow \mathbb{R}\) is a smooth function on the tangent bundle of \(Q\), known as the Lagrangian function of the system.

For every such system the action functional is defined by
\[ S[q] = \int_{t_0}^{t_1} L(q(t), \dot{q}(t)) \, dt, \]
where \(q : [t_0, t_1] \rightarrow Q\) is a smooth curve in \(Q\) and \(\dot{q} : [t_0, t_1] \rightarrow TQ\) is its velocity. An infinitesimal variation of \(q\) is a smooth curve \(\delta q : [t_0, t_1] \rightarrow TQ\) such that $\delta q(t)\in T_{q(t)}Q$ for every $t\in [t_0,t_1]$. An infinitesimal variation is said to have vanishing endpoints if \(\delta q(t_0) = 0\) and \(\delta q(t_1) = 0\).

The dynamics of a Lagrangian mechanical system is determined by Hamilton's Principle, which states that a curve \(q : [t_0, t_1] \rightarrow Q\) is a trajectory of \((Q, L)\) if \(q\) is a critical point of \(S\) for infinitesimal variations \(\delta q\) of \(q\) with vanishing endpoints; that is, \(\mathrm{d}S[q] = 0\) for all infinitesimal variations \(\delta q\) of \(q\) with vanishing endpoints.

This principle gives rise to a set of equations known as the Euler-Lagrange equations for the system \((Q, L)\). Thus, given a set of generalized coordinates $q=(q^i)$ of the configuration space $Q$, a curve \(q: [t_0, t_1] \rightarrow Q\) is a solution of \((Q, L)\) if and only if
\begin{equation}\label{eq:E-L}
     \frac{\mathrm{d}}{\mathrm{d}t} \frac{\partial L}{\partial \dot{q}^i} - \frac{\partial L}{\partial q^i} = 0,\qquad i=1,\dots,n.
\end{equation}
This is a system of \(n\) second-order ordinary differential equations, which is often challenging to solve analytically.

\section{Lagrangian neural networks}

As explained in the previous section, the standard modeling of mechanical systems relies on the assumption that the Lagrangian function $L$ is explicitly known. From this expression, one can symbolically derive the Euler–Lagrange equations to obtain the system's equations of motion. However, as demonstrated by Cranmer and collaborators in \cite{Cranmer2020LagrangianNN}, it is possible to treat $L$ as a black box, thereby allowing the modeling of the system’s dynamics without requiring an analytical expression for the Lagrangian or its associated equations of motion. Within this framework, the Lagrangian formalism still permits a numerical approximation of the system's dynamics. By rewriting the Euler–Lagrange equations in vectorized form and solving for the accelerations, one can compute $\ddot{q}$ from input data $(q,\dot{q})$, even when the Lagrangian is implicitly represented by a neural network. This approach forms the basis for data-driven methods in Lagrangian mechanics, enabling the estimation of physically consistent dynamics directly from observed trajectories.

In \cite{Cranmer2020LagrangianNN}, the authors proposed to derive numerical expressions for the dynamics of Lagrangian systems and expanding the derivatives of the black-box Lagrangian to find the dynamics. Specifically, if we denote the generalized coordinates and velocities by \((q, \dot{q})\), we can write Euler-Lagrange equations \eqref{eq:E-L} in vectorized form as follows:
\begin{equation}
    \frac{\mathrm{d}}{\mathrm{d}t} \nabla_{\dot{q}} L = \nabla_{q} L .
\end{equation}  
where \(\displaystyle\ (\nabla_{\dot{q}})_i = \frac{\partial}{\partial \dot{q}^i}\ \) and \(\ \displaystyle(\nabla_{q})_i = \frac{\partial}{\partial q^i}\ \) are the components of vectors $\nabla_{\dot{q}}$ and $\nabla_{q}$, respectively. Expanding the time derivative and denoting $\nabla_{\dot{q}}^T$ the transpose of the vector $\nabla_{\dot{q}}$, we obtain the expression:
\begin{equation}\label{eq:E-Lvector}
    (\nabla_{\dot{q}} \nabla_{\dot{q}}^T L) \ddot{q} + (\nabla_{q} \nabla_{\dot{q}}^T L) \dot{q} = \nabla_{q} L,
\end{equation} 
where we recognize the products of nabla operators as matrices such that \(\ \displaystyle(\nabla_{\dot{q}} \nabla_{\dot{q}}^T L)_{ij} = \frac{\partial^2 L}{\partial \dot{q}^j \partial \dot{q}^i}\ \) and \(\ \displaystyle(\nabla_{q} \nabla_{\dot{q}}^T L)_{ij} = \frac{\partial^2 L}{\partial q^j \partial \dot{q}^i}\). 

If the Lagrangian is regular, meaning that the matrix \(\ \displaystyle\left(\frac{\partial^2 L}{\partial \dot{q}^i\partial\dot{q}^j}\right)\ \) is invertible, we can solve equation (\ref{eq:E-Lvector}) for the accelerations in terms of the unknown Lagrangian as

\begin{equation}\label{eq:LNN}
\ddot{q} = (\nabla_{\dot{q}} \nabla_{\dot{q}}^T L)^{-1} \left[ \nabla_{q} L - (\nabla_{q} \nabla_{\dot{q}}^T L) \dot{q} \right].
\end{equation}

Thus, given a data set of coordinate pairs $(q,\dot{q})$, we have a procedure for computing the accelerations $\ddot{q}$ from a black‑box Lagrangian.
Using the chain rule for backpropagation process, we take the required derivatives of the network‑estimated Lagrangian and solve the vectorized Euler–Lagrange equations for the accelerations according to equation (\ref{eq:LNN}). 
We then define a loss function as the mean‑squared error (MSE) between these predicted a\-cce\-le\-ra\-tions and the ground‑truth accelerations in the training data. By minimizing this loss, the network learns a Lagrangian that best reproduces the observed dynamics.

\section{Adding constraints}

Lagrangian neural networks have been successfully applied to a wide range of mechanical systems, including unconstrained systems, systems with holonomic constraints (as discussed in \cite{Simplifying} and \cite{ZhongBenchmarking}), and even externally actuated systems, such as in \cite{Leofernandez2021learning} and  \cite{GralizedLNN}. In contrast, our work focuses on systems without external forcing, but with nonholonomic constraints.
In this paper, we develop a different extension of the LNN framework by incorporating nonholonomic constraints directly into the learning process. These constraints depend on both positions and velocities in a nontrivial way, meaning they cannot be expressed solely in terms of positions, as is the case with holonomic constraints.

\vspace{.2cm}
If the system $(Q,L)$ includes nonholonomic constraints, these can be expressed as the common zero set of \( r \) functionally independent functions \(\Phi^{a}:TQ \rightarrow \mathbb{R}\), where \( a = 1, \dots, r \); which can be assembled into an \( r \)-dimensional vector function \(\Phi(q, \dot{q})\). Therefore, the nonholonomic constraints are represented by the \( r \) equations \(\Phi^1(q, \dot{q})=\dots=\Phi^r(q, \dot{q}) = 0\).
We then look for curves \(q: [t_0, t_1] \rightarrow Q\)  such that \(\Phi^a(q(t),\dot q(t)) = 0\) for all \( t \in [t_0, t_1] \) and all \( a = 1, \dots, r \).

\vspace{.2cm}
There are two main approaches for dealing with this case in which the constraints involve velocities in a non-trivial way: the nonholonomic method and the vakonomic method (see, for instance, \cite{Corts2000GeometricDO}, \cite{DELEON2000126} and \cite{LEWIS1995793}). 

\subsection{The vakonomic method}

In the vakonomic setting, a curve $q(t)$ is a trajectory of the system if and only if there are functions $\lambda_a:[t_0,t_1]\to \mathbb R$ such that the curve $(q(t),\lambda(t))$ is stationary for the action corresponding to the augmented Lagrangian $\mathcal{L}$  given by $\mathcal{L}(q,\dot q,\lambda)=L(q,\dot{q})-\Phi(q,\dot{q})^T\lambda$ where $\lambda$ is the vector function $\lambda=(\lambda_1,\dots,\lambda_r)$. That is $\delta S=0$ being  

\begin{equation}\label{accionAumentada}
    \displaystyle S[q,\lambda]=\int \mathcal{L}(q(t),\dot{q}(t),\lambda(t))\;\mathrm{d}t=\int L(q(t),\dot{q}(t))-\Phi(q(t),\dot{q}(t))^T\lambda(t)\;\mathrm{d}t.
\end{equation}

The functions $\lambda_a(t)$ are the so-called Lagrange multipliers and they are introduced as new dynamical variables. 
Enforcing \( \delta S = 0 \) with the action \( S \) as expressed in (\ref{accionAumentada}) yields a system of differential equations that describe the dynamics of the vakonomic system. Variations with respect to Lagrange multipliers \(\lambda\) lead to constraint equations \( \Phi(q, \dot{q}) = 0 \). Meanwhile, variations with respect to \( q \) give us the equation:
\[ \frac{\mathrm{d}}{\mathrm{d}t}(\nabla_{\dot{q}}\mathcal{L}) = \nabla_q \mathcal{L}. \]

This is a set of ordinary differential equations of second order in positions and first order in Lagrange multipliers, which depends on \(\Phi \) and \( L \). Observe that LNNs could still be utilized for the augmented Lagrangian $\mathcal{L}$, as this can be seen as an unconstrained system for that Lagrangian. However, augmented Lagrangians are always singular, a kind of Lagrangian that was not treated in the examples considered in \cite{Cranmer2020LagrangianNN}. On the other hand and from the point of view of implementation, it may be difficult to collect data about Lagrange multipliers to train a LNN model. For these reasons, the case of vakonomic constraints requires special treatment and could be considered in a future work.

\subsection{The nonholonomic method}

In the nonholonomic approach, the system of differential equations describing the dynamics of the system may be also derived from a variational-like principle as follows. A curve $q(t)$ is a trajectory of the nonholonomic system if and only if the following equations are satisfied (summation over repeated indices is assumed from now on):

\begin{eqnarray*}
\frac{\mathrm{d}}{\mathrm{d}t}\frac{\partial L}{\partial \dot{q}} - \frac{\partial L}{\partial q} & = & \lambda_a \displaystyle\frac{\partial \Phi^a}{\partial \dot{q}}\\
\Phi^a(q, \dot{q})  & = & 0,
\end{eqnarray*}
where the functions \(\lambda_a(t)\) are unknowns that must also be determined. Similar to the vakonomic method, these new variables are called Lagrange multipliers. Nonetheless, we stress that they are not the same multipliers as in vakonomic case and, in fact, both systems of differential equations give rise to different trajectories except in case of holonomic constraints \cite{LEWIS1995793}.

By expanding the total time derivative, we obtain the equations of motion as follows:

$$\frac{\partial^2 L}{\partial \dot{q}^j \partial \dot{q}^i} \ddot{q}^j + \frac{\partial^2 L}{\partial q^j \partial \dot{q}^i} \dot{q}^j = \frac{\partial L}{\partial q^i} + \lambda_a \frac{\partial \Phi^a}{\partial \dot{q}^i} $$
or, equivalently, 
\begin{equation}\label{AcelerMultip}
    \frac{\partial^2 L}{\partial \dot{q}^j \partial \dot{q}^i} \ddot{q}^j = \frac{\partial L}{\partial q^i} + \lambda_a \frac{\partial \Phi^a}{\partial \dot{q}^i} - \frac{\partial^2 L}{\partial q^j \partial \dot{q}^i} \dot{q}^j\ .
\end{equation}

\subsection{Nonholonomic Lagrangian neural networks}

We recall that the central question of this work is whether a tailored adaptation of Lagrangian neural networks can more effectively accommodate nonholonomic constraints. To address this, we propose modifying the procedure used to compute accelerations during the learning process of the LNN so as to explicitly incorporate the constraints when present in the mechanical system. The loss function is then defined using the appropriate expression for the accelerations dictated by nonholonomic dynamics.

\vspace{.2cm}

For the purpose of learning a regular Lagrangian, and assuming the constraints are known, the above formulation allows us to express the accelerations 
\(\ddot{q}\) in terms of the Lagrangian \(L\), the constraints \(\Phi^a\), and the Lagrange multipliers \(\lambda_a\), in a manner analogous to the unconstrained case, as follows.

\vspace{.2cm}
Using the vectorized nabla symbol, we can write equation (\ref{AcelerMultip}) as:
\begin{align*}
\nabla_{\dot{q}}\nabla_{\dot{q}}^TL\cdot\ddot q&=\nabla_q L+\lambda_a \nabla_{\dot q}\Phi^a-\nabla_{\dot q}\nabla_q L\cdot\dot q
\end{align*}
and, hence, for nonsingular Lagrangians, the accelerations are given by:
\begin{align}\label{acelerations}
\ddot q&=\left[\nabla_{\dot{q}}\nabla_{\dot{q}}^TL\right]^{-1}\cdot\left[\nabla_q L+\lambda_a \nabla_{\dot q}\Phi^a-\nabla_{\dot q}\nabla_q L\cdot\dot q\right].
\end{align}

As we remarked before, the Lagrange multipliers \(\lambda_a\) are also unknowns. However, if a given curve satisfies the constraints \(\ \Phi^a(q(t), \dot{q}(t)) = 0,\ \) then we can differentiate with respect to time to find $$\displaystyle 0=\frac{\mathrm{d}\Phi}{\mathrm{d}t}=\frac{\partial\Phi^a}{\partial q^i}\dot q^i+\frac{\partial\Phi^a}{\partial \dot q^i}\ddot q^i\quad\quad \text{ and then }\quad\quad \displaystyle\frac{\partial\Phi^a}{\partial \dot q^i}\ddot q^i=-\frac{\partial\Phi^a}{\partial q^i}\dot q^i.$$ Using the equations of motion (\ref{acelerations}), the last equality can be rewritten in matrix notation as follows:
$$
\nabla_{\dot q}\Phi^a\cdot \left(\left[\nabla_{\dot{q}}\nabla_{\dot{q}}^TL\right]^{-1}\cdot\left[\nabla_q L+\lambda_b \nabla_{\dot q}\Phi^b-\nabla_{\dot q}\nabla_q L\cdot\dot q\right]\right)=-\nabla_q\Phi^a\cdot\dot q$$ or equivalently,
$$\nabla_{\dot q}\Phi^a\cdot \left[\nabla_{\dot{q}}\nabla_{\dot{q}}^TL\right]^{-1}\cdot \nabla_{\dot q}\Phi^b \lambda_b =-\nabla_q\Phi^a\cdot\dot q-\nabla_{\dot q}\Phi^a\cdot\left[\nabla_{\dot{q}}\nabla_{\dot{q}}^TL\right]^{-1}\cdot\left[\nabla_q L-\nabla_{\dot q}\nabla_q L\cdot\dot q\right]$$

\vspace{.3cm}
To find a compact form of this expression, we can define the force \( f \) as:

\[ f = \nabla_{q} L - \nabla_{\dot{q}} \nabla_{q} L \cdot \dot{q} \]
and define an \( r \times r \) matrix \( M \) whose entries are given by
\[ M^{ab} = \nabla_{\dot{q}} \Phi^a \cdot \left[\nabla_{\dot{q}} \nabla_{\dot{q}}^T L\right]^{-1} \cdot \nabla_{\dot{q}} \Phi^b . \]

Then, inverting \( M \) whenever possible, we can solve the last equation for the Lagrange multipliers
\begin{align*}
\lambda_b &=-M^{-1}_{ba}\left(\nabla_q\Phi^a\cdot\dot q+\nabla_{\dot q}\Phi^a\cdot\left[\nabla_{\dot{q}}\nabla_{\dot{q}}^TL\right]^{-1}\cdot f\right).
\end{align*}

Gathering all this together and denoting \(\ \nabla_{\dot{q}} \nabla_{\dot{q}}^T = \nabla^2_{\dot{q}}\ \), the equation of motion that must be considered along with the constraint equation $\ \Phi^a(q, \dot{q}) = 0,\ $ can be written as follows:
\begin{align}\label{aceleracionesnh}
\ddot q&=\left[\nabla^2_{\dot q}L\right]^{-1}\cdot\left[f- \nabla_{\dot q}\Phi^b M^{-1}_{ba}\left(\nabla_q\Phi^a\cdot\dot q+\nabla_{\dot q}\Phi^a\cdot\left[\nabla^2_{\dot q}L\right]^{-1}\cdot f\right)\right].
\end{align}

This equation can be seen as a matrix version of equations in  \cite{ManuelDavidJCarlos} and a generalization of equations obtained in \cite{Simplifying} and \cite{LaValle} to the case where constraints depend on velocities in a non trivial manner.

In the remaining sections we will use Eq. \eqref{aceleracionesnh} instead of Eq. \eqref{eq:LNN} to train a Lagrangian neural network capturing the nonholonomic constrained nature of the dynamics of the system (including holonomic constraints as a particular case). Note that equation \eqref{eq:LNN} can be recovered from \eqref{aceleracionesnh} in the absence of constraints. 

To easily distinguish from the original LNN, we will refer to these Lagrangian neural networks for nonholonomic systems as LNN-nh. 

In three of the examples we implement, the constraint happens to be linear, so we dedicate the next subsection to obtain a simpler expression of the previous equations for this type of restriction.

\subsubsection{Linear constraints}\label{sec:linear}

If we consider the case in which the constraints \(\Phi^a(q, \dot{q})\) are given by differential 1-forms \(\omega^a\) such that \(\Phi^a(q, \dot{q}) = \omega^a(q) \cdot \dot{q}\), the system of equations of motion is given by:
$$ \left\{ \begin{array}{rl} 
\displaystyle\frac{\mathrm{d}}{\mathrm{d}t}\frac{\partial L}{\partial\dot{q}}-\frac{\partial L}{\partial q} &=\lambda(t)\omega(q) \\
         \omega(q)\cdot\dot{q} &=0
         \end{array}\right..$$
In this case, we can write $\nabla_q\Phi^a=\nabla_q\omega^a\cdot \dot q\ ,\ \nabla_{\dot q}\Phi^a=\omega^a$ and $ 
M^{ab}=\omega^a\cdot \left[\nabla_{\dot{q}}^2L\right]^{-1}\cdot\omega^b$.

Therefore 
\begin{align*}
\ddot q&=\left[\nabla_{\dot{q}}^2L\right]^{-1}\cdot\left[f-\omega^b M^{-1}_{ba}\left(\dot q\cdot\nabla_q\omega^a\cdot\dot q+\omega^a\cdot\left[\nabla_{\dot{q}}^2L\right]^{-1}\cdot f\right)\right]
\end{align*}
and we can express the Lagrange multipliers as
$$
\lambda_b =-M^{-1}_{ba}\left(\dot q\cdot\nabla_q\omega^a\cdot\dot q+\omega^a\cdot\left[\nabla_{\dot{q}}^2L\right]^{-1}\cdot f\right).
$$

As a result, we have the accelerations for a system with linear nonholonomic constraints written as
\begin{align*}
\ddot q&=\left[\nabla_{\dot{q}}^2L\right]^{-1}\cdot\left[f+\omega^b\lambda_b\right].
\end{align*}

\section{Implementation}

We evaluate our proposed method by applying it to four representative systems: a nonholonomic particle, a rolling wheel, a dog chasing a man and a particle subjected to a constant gravitational field. The first three examples present linear nonholonomic constraints, so we follow the calculations described in Section \ref{sec:linear} for these cases. We implement the neural networks similarly to the approach developed in \cite{Cranmer2020LagrangianNN}, as detailed in the subsequent subsection.

\subsection{Datasets}\label{sec:datasets}

We generate data for each system by taking 500 \emph{true} trajectories as follows: first, we sampled 500 initial states of the form $(q,\dot q)$ fulfilling the constraint. The initial states for all examples were sampled using an uniform distribution on a fixed range (minimum and maximum values of the range depend on the specific example). Then, we simulate 500 different trajectories starting at the mentioned initial states for 1000 timesteps (ranging from 0 to 10) using the true equations of motion (that is, the equations derived analytically using the true known Lagrangian of the system). Thus, we end up with 500,000 pairs of the form $(q,\dot q)$.

This simulation was made with an implementation of an adaptive-step Dormand-Prince method in the examples with linear constraints, so it is a numerical simulation, but we refer to these trajectories as \emph{true} trajectories because they were generated using the true Lagrangian. In the example of the particle subjected to a constant gravitational field, equations can be solved analytically, so we were able to simulate trajectories without the use of any numerical integrator.  

Finally, for every pair $(q,\dot q)$ in each dataset, acceleration targets $\ddot q_{true}$ are generated using the true Lagrangian and the constraint function together with Eq. \eqref{aceleracionesnh}, for LNN-nh model, and the true Lagrangian together with Eq. \eqref{eq:LNN} for LNN model.

\subsection{Training details and architecture}
One of our objectives is to compare the performance of our LNN-nh approach, which considers nonholonomic constraints, with that of Lagrangian Neural Networks in the mentioned examples. To achieve this, we implement and train our models using JAX Python library, following the methodology outlined in \cite{Cranmer2020LagrangianNN}.

\vspace{.2cm}

For each model (LNN and LNN-nh), we use a four-layer MLP architecture. The neural network will be trained to learn the Lagrangian function of each example. Thus the output layer will have a single unit and the input layer must have as many units as the number of positions and velocities of the system. We choose the two hidden layers to contain 128 fully connected neurons each. We employ softplus for activations and use ADAM optimizer with adaptive learning rates starting at $10^{-3}$. In each example, we follow a stochastic gradient descent strategy, training the models for 300 epochs. During each epoch we generate a dataset of 500 \emph{true} trajectories fulfilling the constraint (as detailed in the previous subsection). We then take minibatches of size 1000, so that each minibatch corresponds to a complete trajectory of the dataset generated in the current epoch. 
We divide the dataset of each epoch, using 88\% for training and the remaining 12\% for testing (that is the last 60 trajectories of every epoch).

All the implementation was carried out using JAX libraries \quad \texttt{jax.example\_libraries.stax} \quad and \newline \texttt{jax.example\_libraries.optimizers}.

\subsection{Performance}

In each example, we evaluate the performance of both models by considering the dynamics obtained from the prediction of the Lagrangian learned from LNN and from LNN-nh models. We compare loss function, predictions, energy and constraint from both models.

Loss function is taken to be the mean squared error in accelerations, as said before. More specifically, for each pair $(q,\dot q)$ and corresponding target $\ddot q_{true}$ we compute the predicted $\ddot q_{pred}$ using the Lagrangian learned with the LNN-nh model and the constraint together with Eq. \eqref{aceleracionesnh}, and the Lagrangian learned with the LNN model together with Eq. \eqref{eq:LNN}. Then we take the MSE of the difference $\ddot q_{true}-\ddot q_{pred}$ and sum and normalize over all the minibatch. 

For energy and constraint assessment, we simulate five different trajectories for each model (ten trajectories total) following the process outlined in subsection \ref{sec:datasets} but this time using the Lagrangians learned by the models instead of the true Lagrangian (we call these trajectories \emph{learned} trajectories to distinguish them from the true trajectories). We stress that these trajectories are not part of the training nor the testing set, since they are simulated using the models. We then compute the true energy and constraint along these learned trajectories.

\section{Examples}

In this section, we illustrate the performance of our proposed method across four representative mechanical systems subjected to nonholonomic constraints. The first three examples involve systems with linear ve\-lo\-ci\-ty constraints, while the fourth example presents a nonlinear nonholonomic constraint.
These examples illustrate how the proposed adaptation of Lagrangian neural networks performs when learning the dynamics of systems subjected to different types of nonholonomic constraints. For each system, we compare the LNN-nh model with a standard LNN by analyzing the loss function, the predicted accelerations and evaluating how well each approach preserves energy and satisfies the constraints along learned trajectories.
The results suggest that incorporating the constraint structure during training can lead to improved performance in capturing the system’s dynamics.

\subsection{The nonholonomic particle}

Consider a free particle of mass 1 moving in three-dimensional space with standard coordinates in $\mathbb R^3$, subjected to the nonholonomic constraint given by $\Phi(q,\dot q)=\Phi(x,y,z,\dot{x},\dot{y},\dot{z})=\dot z-y\dot x=0$. Notice that the constraint $\Phi$ is linear in velocities since $\Phi(q,\dot q)=\omega(q)\cdot\dot{q}$ with the differential 1-form $\omega(q)=-ydx+dz,$ that is $(-y,0,1)$ in coordinates.

The Lagrangian of the system is given by $\ L(q,\dot q)=L(x,y,z,\dot{x},\dot{y},\dot{z})=\displaystyle \frac{1}{2}(\dot x^2+\dot y^2+\dot z^2)\ $ and then, following Section \ref{sec:linear}, we have $\ M=1+y^2\ $ and $\ f=(0,0,0).\ $ Subsequently, we obtain the equation of motion of the nonholonomic particle given by 
$$
\begin{pmatrix}
\ddot x\\
\ddot y\\
\ddot z
\end{pmatrix}
=\frac{\dot x \dot y}{1+y^2}
\begin{pmatrix}
-y\\
0\\
1
\end{pmatrix}.
$$

\begin{figure}[h!]
    \centering
    \includegraphics[width=0.85\linewidth]{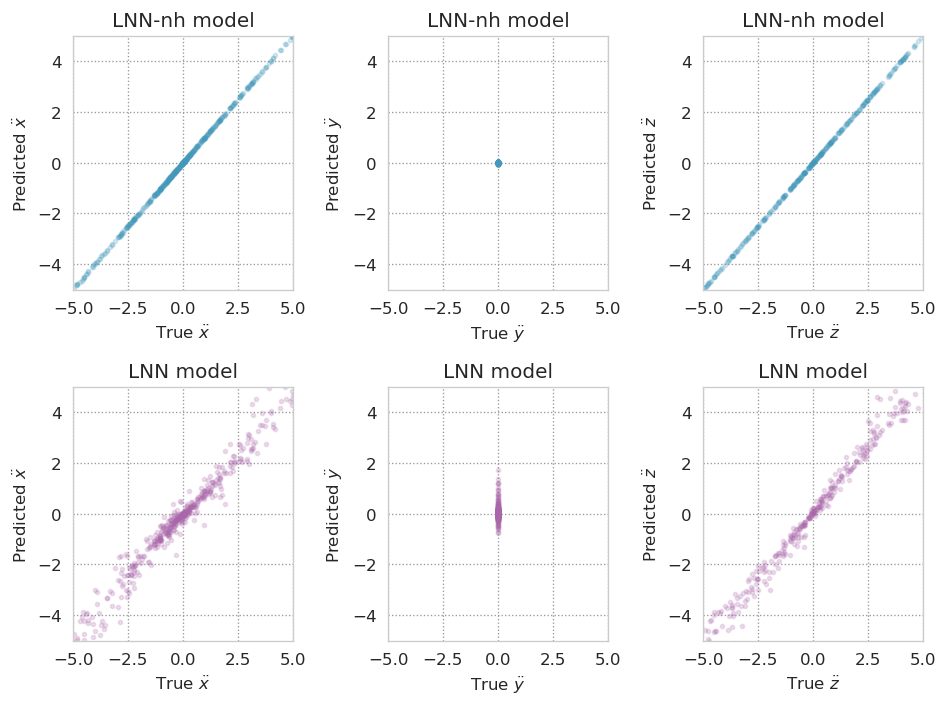}
    \caption{\label{fig:dispersionparticula} The scatter plots show five hundred true vs. corresponding learned cartesian accelerations for LNN-nh and LNN models in the nonholonomic particle example.}
\end{figure}

For the example, we randomly selected five hundred pairs $(q,\dot q)$ and computed the corresponding accelerations using both models. The resulting scatter plots, shown in Figure \ref{fig:dispersionparticula}, compare each component of $\ \ddot q_{pred}\ $ and $\ \ddot q_{true}\ $ for the LNN and LNN-nh models. The plots reveal a greater dispersion in the accelerations predicted by the LNN model.

\begin{figure}[h!]
   \begin{subfigure}[b]{0.55\linewidth}
   \centering
   \includegraphics[width=\linewidth]{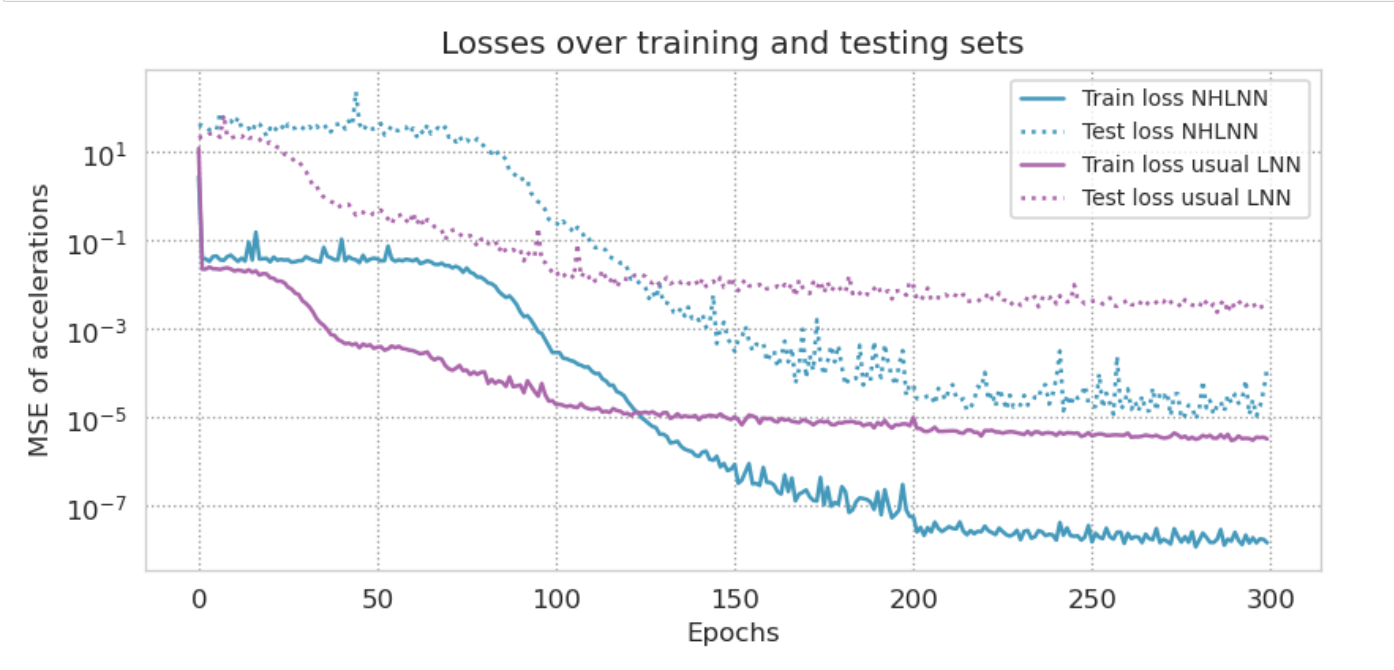}
   \caption{Loss function}
     \label{fig:perdidaparticle}
   \end{subfigure}
\begin{subfigure}[b]{0.4\linewidth}
\includegraphics[width=\linewidth]{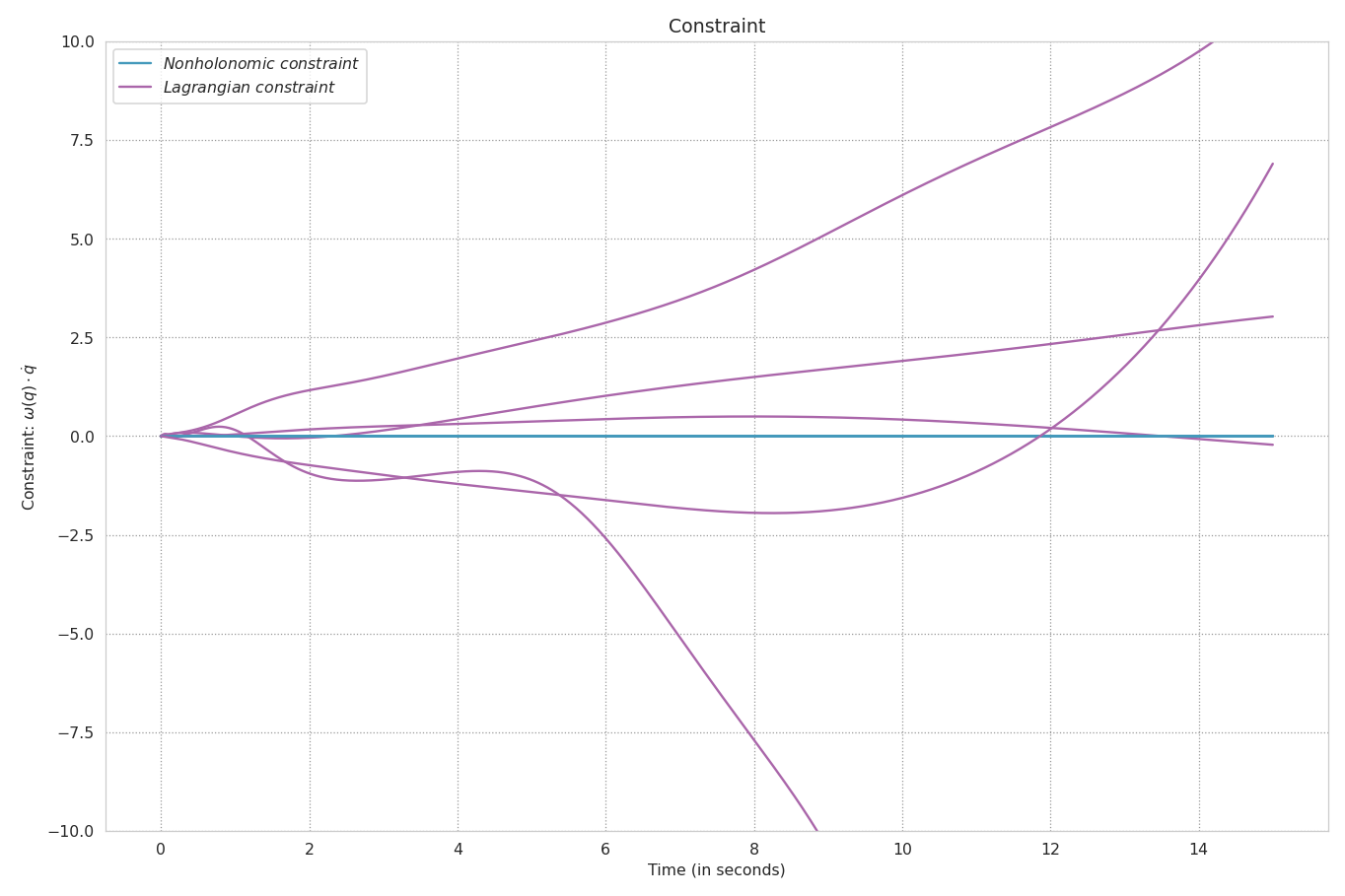}  
\caption{Nonholonomic constraint}
\label{fig:restriccionparticula}
\end{subfigure}
\centering
\begin{subfigure}[b]{0.43\linewidth}
\includegraphics[width=\linewidth]{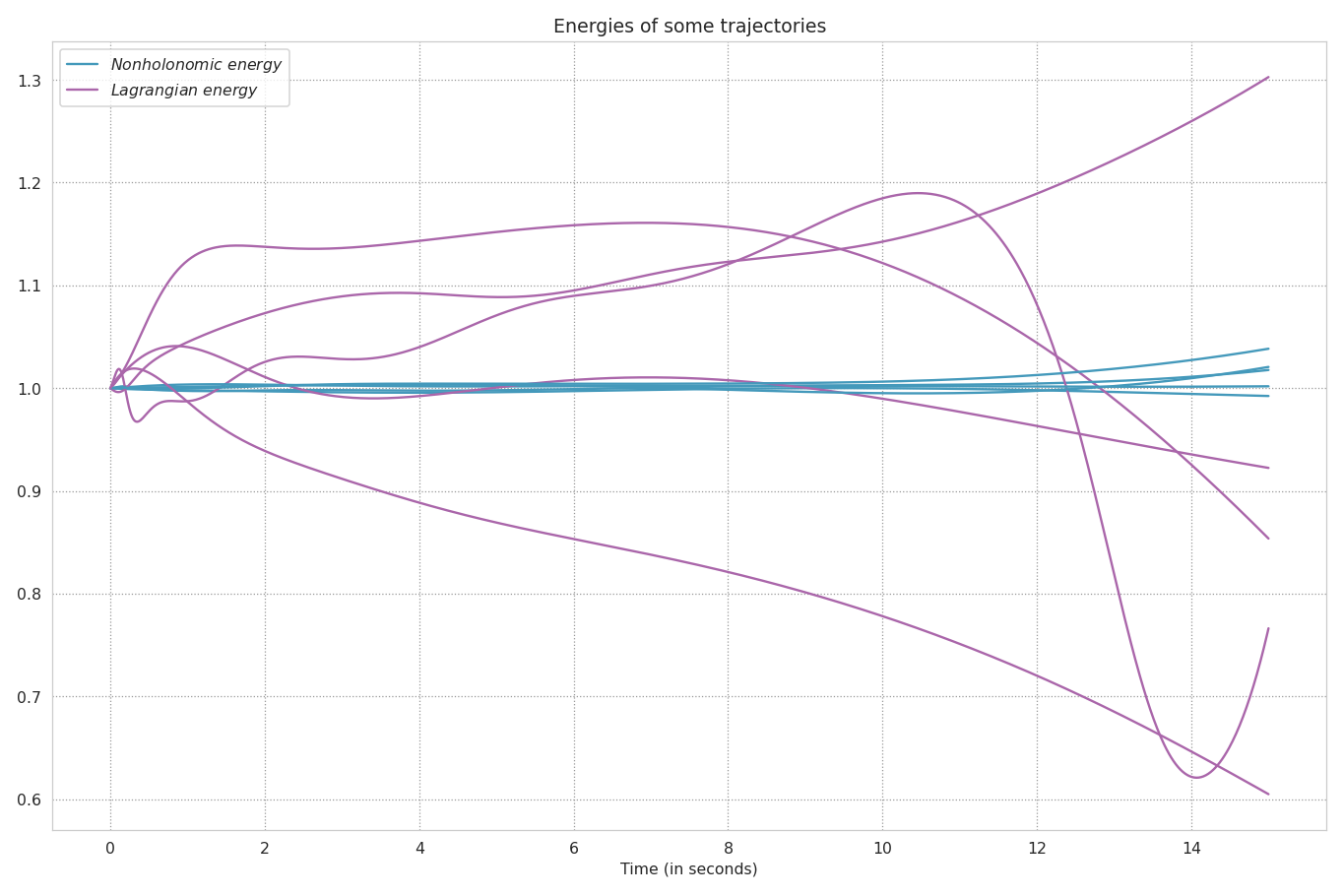}
\caption{Total energy}
\label{fig:energiaparticula}
\end{subfigure}
\begin{subfigure}[b]{0.43\linewidth}
\includegraphics[width=\linewidth]{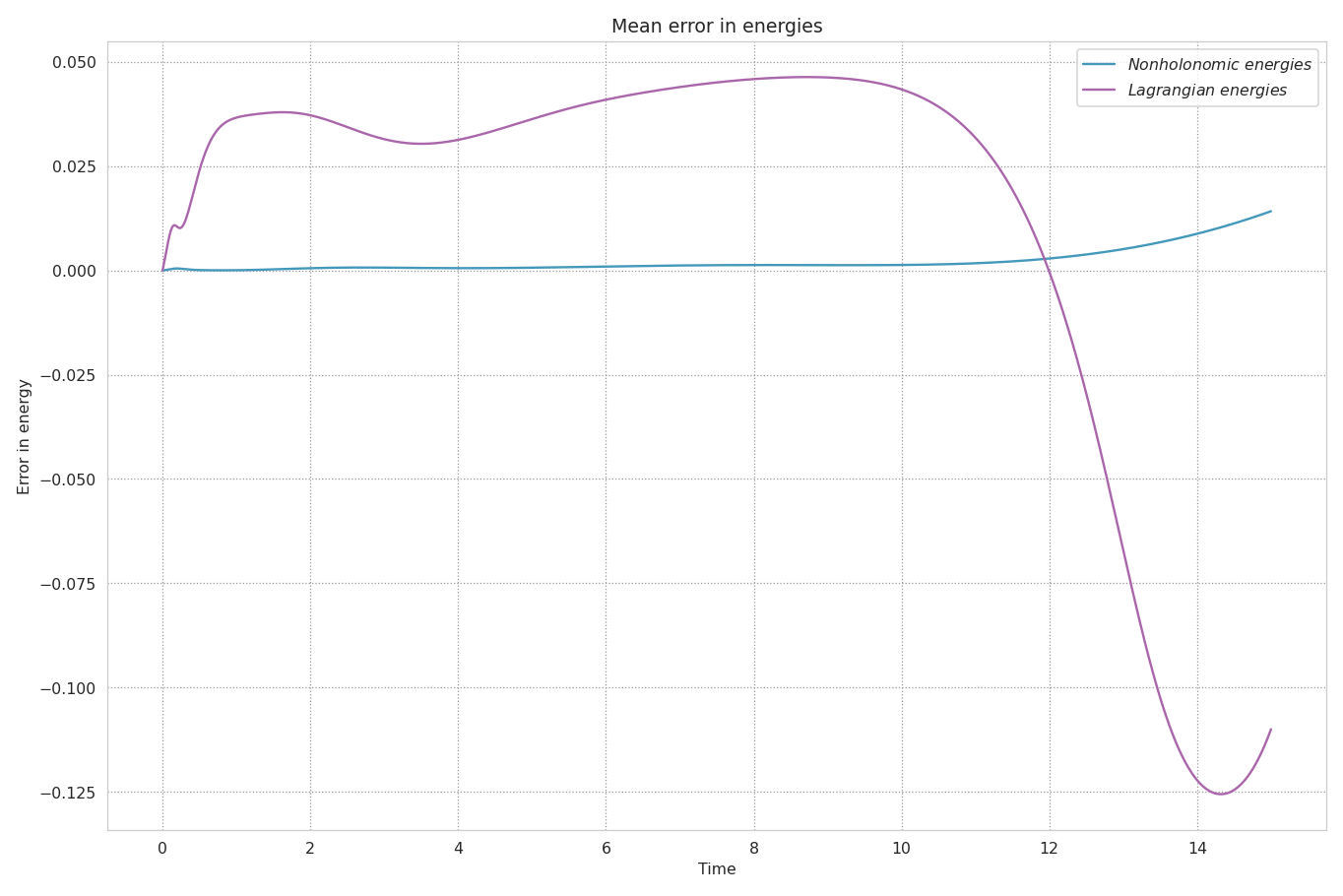}  
\caption{Mean error of energies}
\label{fig:errorparticula}
\end{subfigure}

\caption{\label{fig:particle}(\ref{fig:perdidaparticle}) Comparison of the loss function from the nonholonomic particle for training and testing sets corresponding to LNN and LNN-nh models. Picture (\ref{fig:restriccionparticula}) shows the evolution of the constraint function along five trajectories of the nonholonomic particle generated from the same initial conditions for both models.  Picture (\ref{fig:energiaparticula}) exhibits the total energy of each trajectory normalized with the corresponding constant true energy. Picture (\ref{fig:errorparticula}) shows the mean relative error in energy of the trajectories.} 
\end{figure}

Figure \ref{fig:particle} shows the comparison between the loss function in the case in which we learn the Lagrangian using a LNN and with a LNN-nh. Performance of energy and constraint functions over five learned trajectories of both models are also shown in the same picture.

\subsection{A dog pursuing a man}

Consider a dog and a man moving in the plane. We take the man as a particle of mass $m_t$ moving freely along the $y$ axis and we take the dog as another particle of mass $m_d$ moving in the plane pursuing the man, i.e. with velocity pointing directly to it (see \cite{Swaczyna2011SeveralEO} for details). The position of the target man is determined with a single coordinate $w$, whereas position of the dog may be described with two cartesian coordinates $(x,y)$. Hence, the system's state is completely described by the tuple $(q, \dot q)=(w,x,y,\dot w, \dot x, \dot y)$.

The Lagrangian of this system is given by $$L(q, \dot q)=L(w,x,y,\dot w, \dot x, \dot y)=\frac{1}{2}m_t\dot{w}^2+\frac{1}{2}m_d(\dot{x}^2+\dot{y} ^2)$$ and the single constraint can be written as 
\[
\Phi(w,x,y,\dot{w},\dot{x},\dot{y})=x\dot{y}+(w-y)\dot{x}=0.
\]

\vspace{.5cm}
Accordingly, $\nabla_q\Phi=(\dot{x},\dot{y},-\dot{x})$, so we notice that the restriction is in fact linear, i.e. $\Phi(q,\dot{q})=\omega(q)\cdot \dot q$ with $\omega=\nabla_{\dot{q}}\Phi=(0,w-y,x)$. In this case, the matrix $M$ is the scalar $M=\displaystyle\frac{x^2+(w-y)^2}{m_d}$ and we have a unique Lagrange multiplier given by
\[
\lambda =  -\frac{m_d\dot x \dot w}{x^2+(w-y)^2}.
\]
Similar to the nonholonomic particle, we have no potential, so the force vanishes $f=(0,0,0)$. 

Gathering all this information, we can write the equations of motion as 
$$
\begin{pmatrix}
\ddot w\\
\ddot x\\
\ddot y
\end{pmatrix}
=\frac{\dot x \dot w}{x^2+(w-y)^2}
\begin{pmatrix}
0\\
y-w\\
-x
\end{pmatrix}
$$

\begin{figure}[h!]
    \centering
    \includegraphics[width=0.85\linewidth]{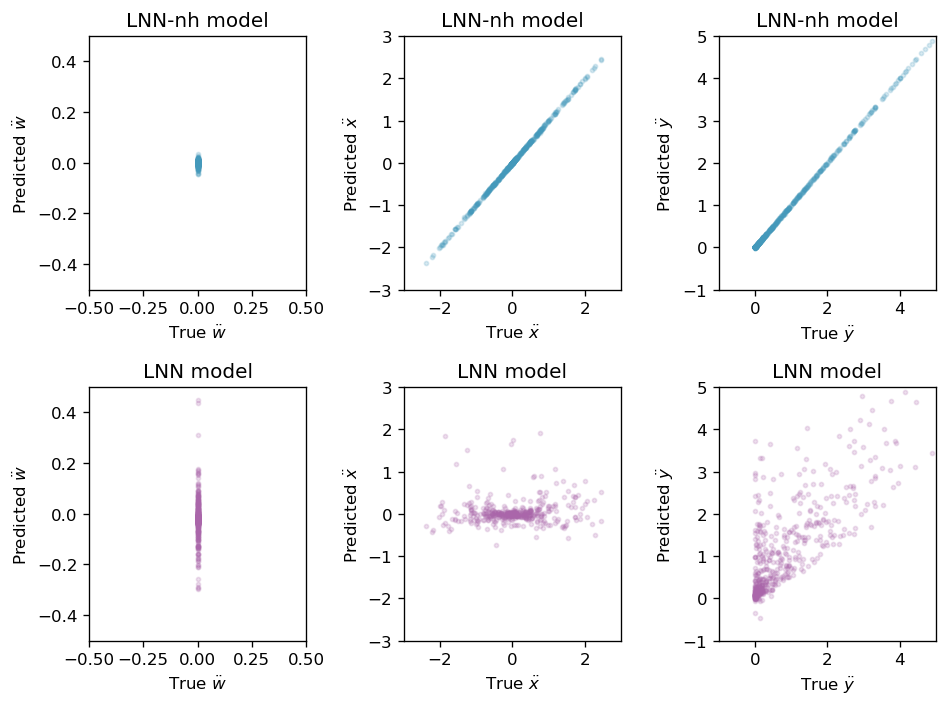}
    \caption{\label{fig:dispersionperrito} The scatter plots show five hundred true vs. corresponding learned cartesian accelerations for LNN-nh and LNN models in the man-dog example.}
\end{figure}

Figure \ref{fig:dispersionperrito} shows the comparison between the learned and true value of each coordinate acceleration in the example for both models, exhibiting a major dispersion in LNN learned accelerations. 

\begin{figure}[h!]
   \begin{subfigure}[b]{0.55\linewidth}
   \centering
   \includegraphics[width=\linewidth]{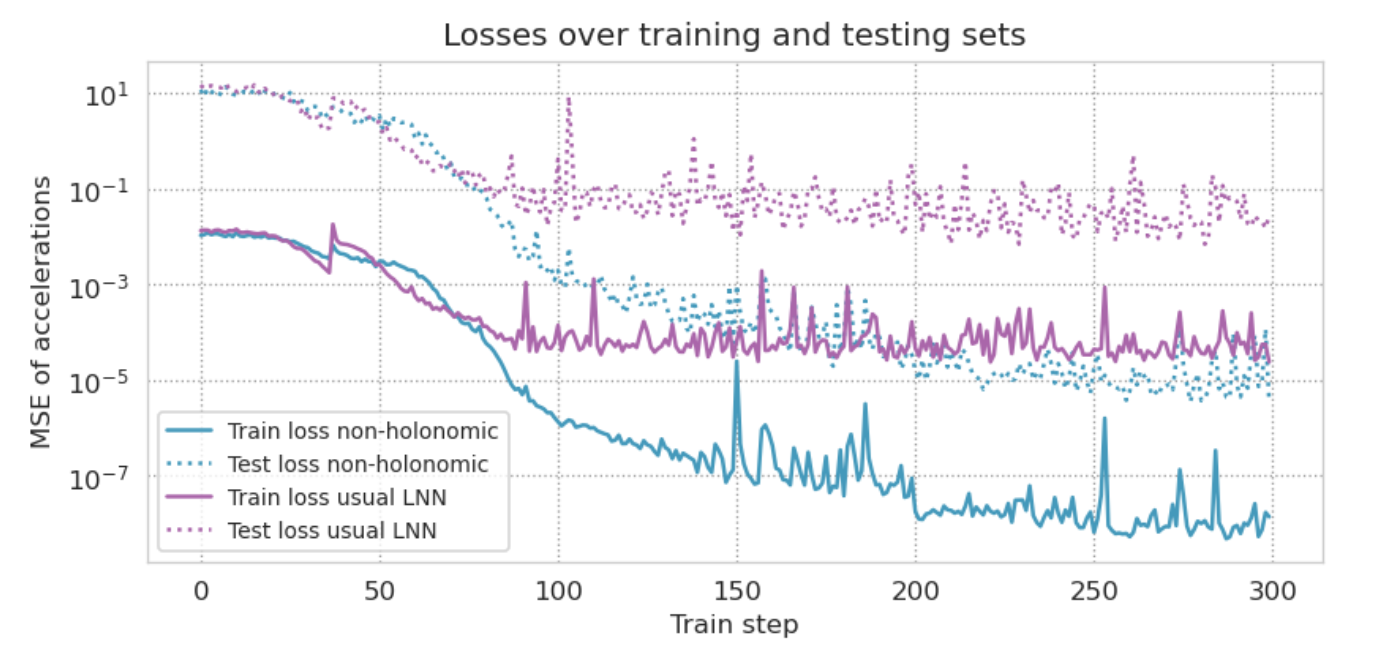}  
   \caption{Loss function}
     \label{fig:perdidaperrito}
   \end{subfigure}
   \begin{subfigure}[b]{0.33\linewidth}
\includegraphics[width=\linewidth]{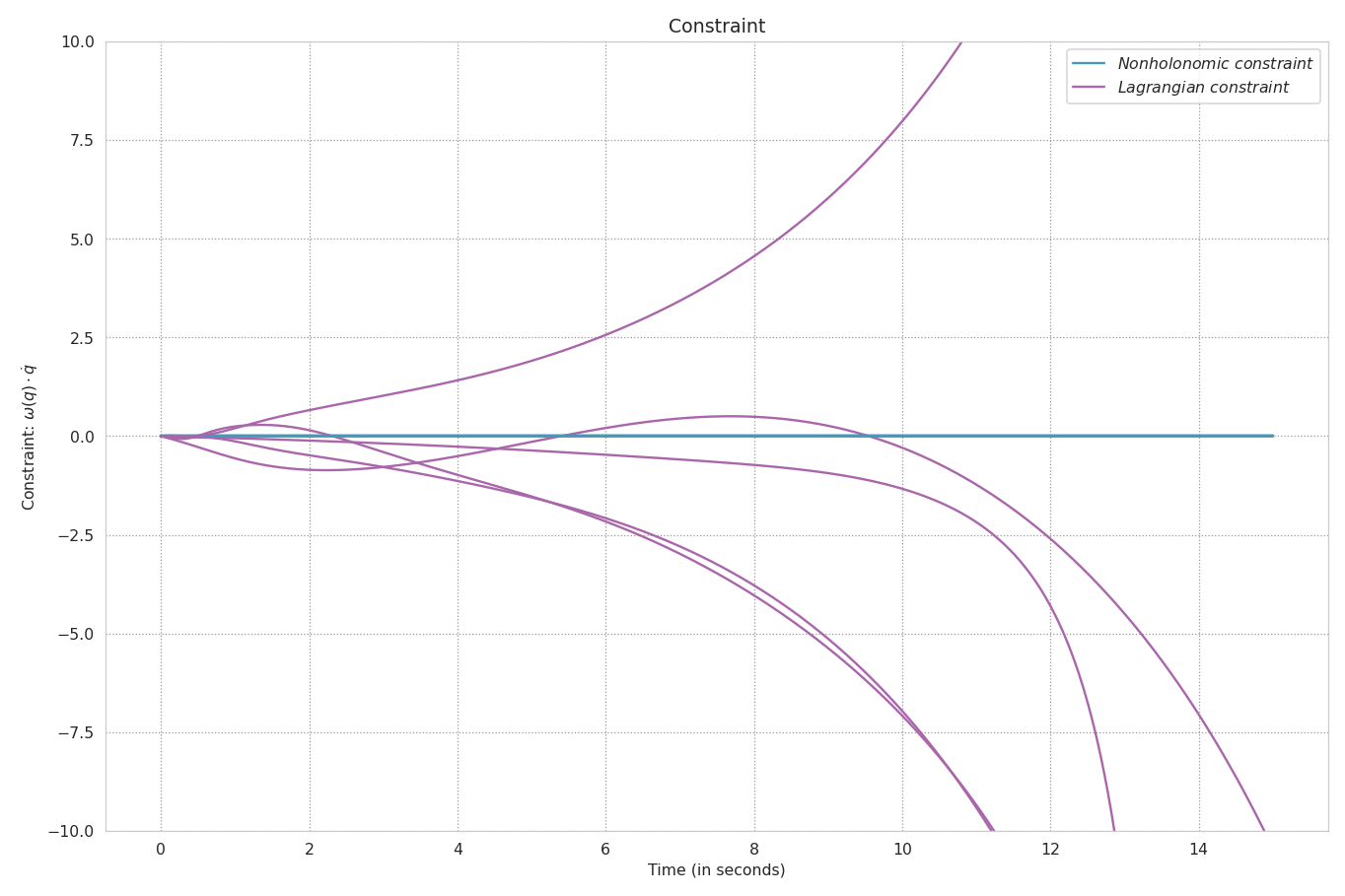}   
\caption{Nonholonomic constraint}
\label{fig:restriccionperrito}
\end{subfigure}
\centering
\begin{subfigure}[b]{0.4\linewidth}
\includegraphics[width=\linewidth]{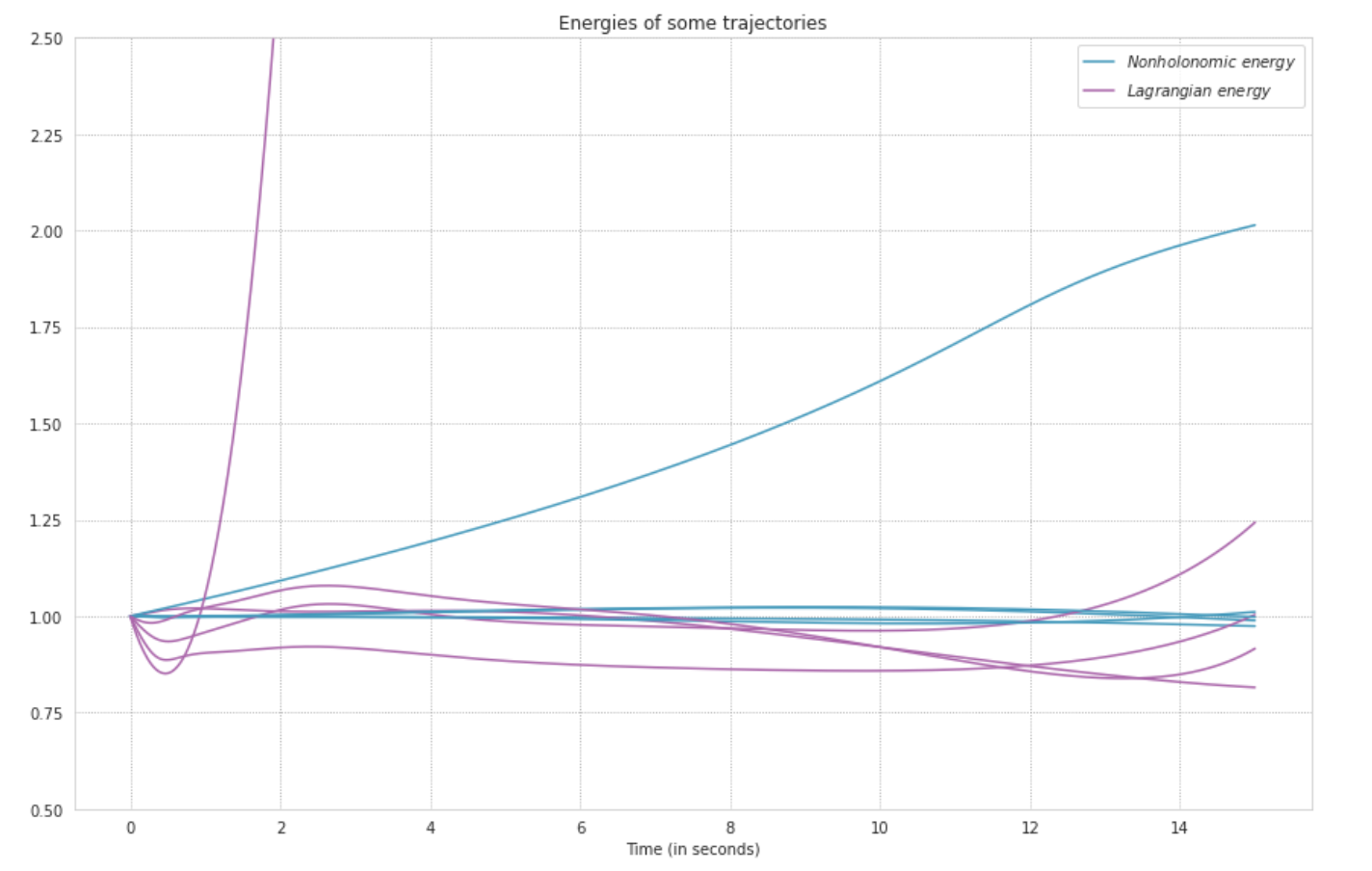} 
\caption{Total energy}
\label{fig:energiaperrito}
\end{subfigure}
\begin{subfigure}[b]{0.4\linewidth}
\includegraphics[width=\linewidth]{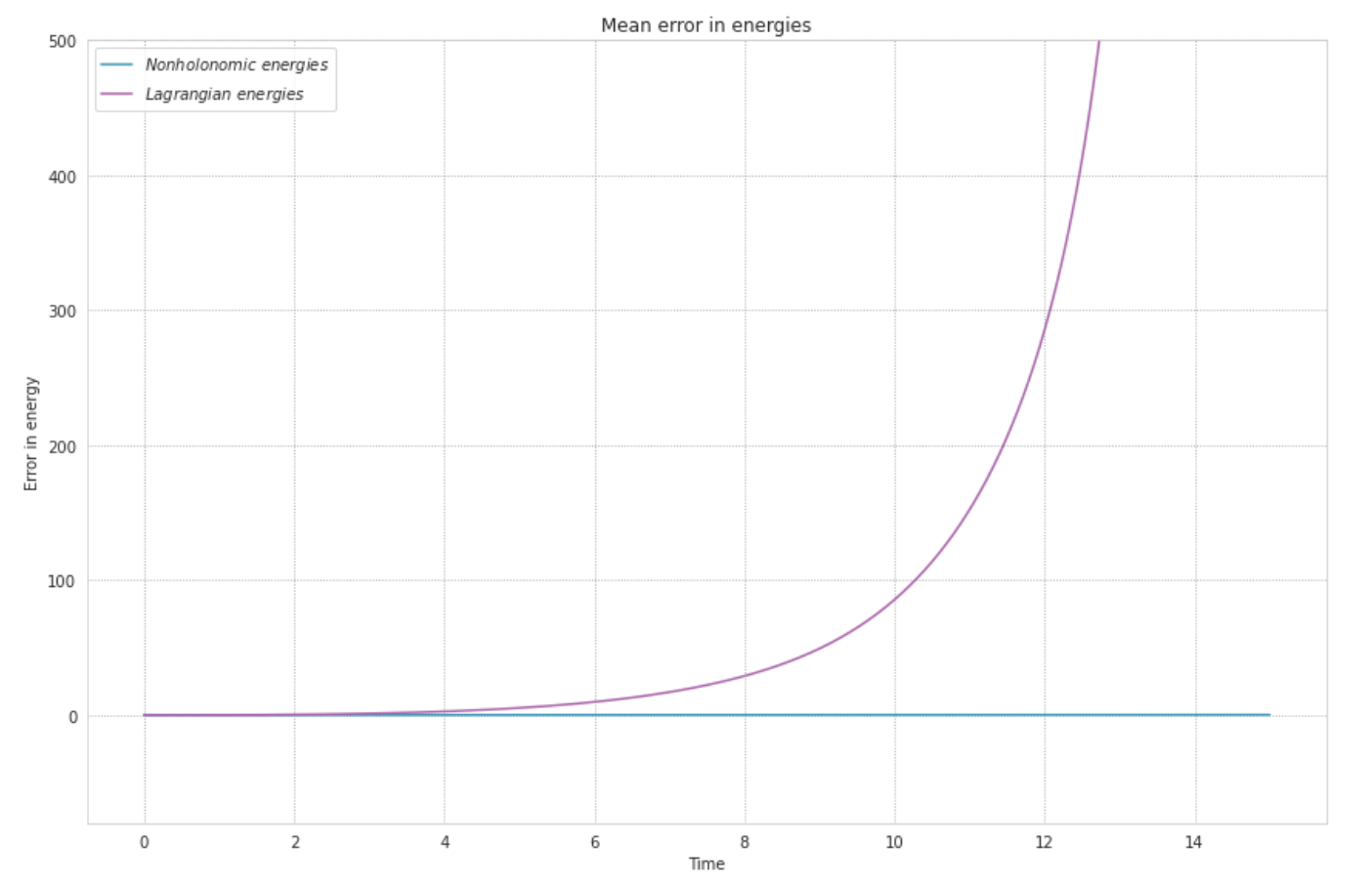}   
\caption{Mean error of energies}
\label{fig:errorperrito}
\end{subfigure}
\caption{\label{fig:perrito} (\ref{fig:perdidaperrito}) Comparison of the loss function from the man-dog example for training and testing sets corresponding to LNN and LNN-nh models. Picture (\ref{fig:restriccionperrito}) shows the evolution of the constraint function along five trajectories of the man-dog system generated from the same initial conditions for both models. Picture (\ref{fig:energiaperrito}) exhibits the total energy of each trajectory normalized with the corresponding constant true energy. Picture (\ref{fig:errorperrito}) shows the mean relative error in energy of the trajectories.}
\end{figure}

In Figure \ref{fig:perrito} we have included the results from models LNN and LNN-nh of the loss function, and the energy and constraint values for five different trajectories of the learned dynamics of the example. Computations are performed considering $m_d=m_t=1$.

\subsection{A vertical rolling wheel} 

A configuration of a wheel as a disk rolling without slipping in a vertical position in a plane is given by a point $q=(x,y,\theta,\phi)\in\ \mathbb{R}^2 \times S^1\times S^1$. The meaning of the variables is detailed, for instance, in \cite{CDplano, baillieul2008nonholonomic}. 

The Lagrangian is given by 
\begin{equation*}
L(q,\dot{q})=L(x,y,\theta,\phi,\dot{x},\dot{y},\dot{\theta},\dot{\phi})=\displaystyle\frac{m}{2}(\dot{x}^2+\dot{y}^2)+\frac{1}{2}I\dot{\theta}^2+\frac{1}{2}J\dot{\phi}^2,    
\end{equation*} where $m$ is the mass of the wheel and $I,\ J$ are the momenta of inertia. We consider $m=1$, $I=0.5$ and $J=0.25$ in implementation. The rolling-without-slipping restriction is a constraint of rank two given by the equations 
\begin{equation*}
\begin{cases}
\Phi^1(x,y,\theta,\phi,\dot{x},\dot{y},\dot{\theta},\dot{\phi})=\dot{x}-R\cos(\phi)\dot{\theta}&=0 \\
\Phi^2(x,y,\theta,\phi,\dot{x},\dot{y},\dot{\theta},\dot{\phi})=\dot{y}-R\sin(\phi)\dot{\theta}&=0.
\end{cases}
\end{equation*} 

As in the previous examples, $f=(0,0,0,0)$. On the other hand, we have $\nabla_{\dot{q}}\Phi^1=(1,0,0,-R\cos(\phi)),$

$\nabla_{\dot{q}}\Phi^2=(0,1,0,-R\sin(\phi))$  and $$M=\begin{pmatrix}
\displaystyle\frac{1}{m}+\frac{R^2}{I}\cos^2(\phi)  & \quad
\displaystyle\frac{R^2}{I}\sin(\phi)\cos(\phi)  \\
\displaystyle\frac{R^2}{I}\sin(\phi)\cos(\phi)  & \quad
\displaystyle\frac{1}{m}+\frac{R^2}{I}\sin^2(\phi) \\
\end{pmatrix}.$$

\begin{figure}[h!]
    \centering
    \includegraphics[width=\linewidth]{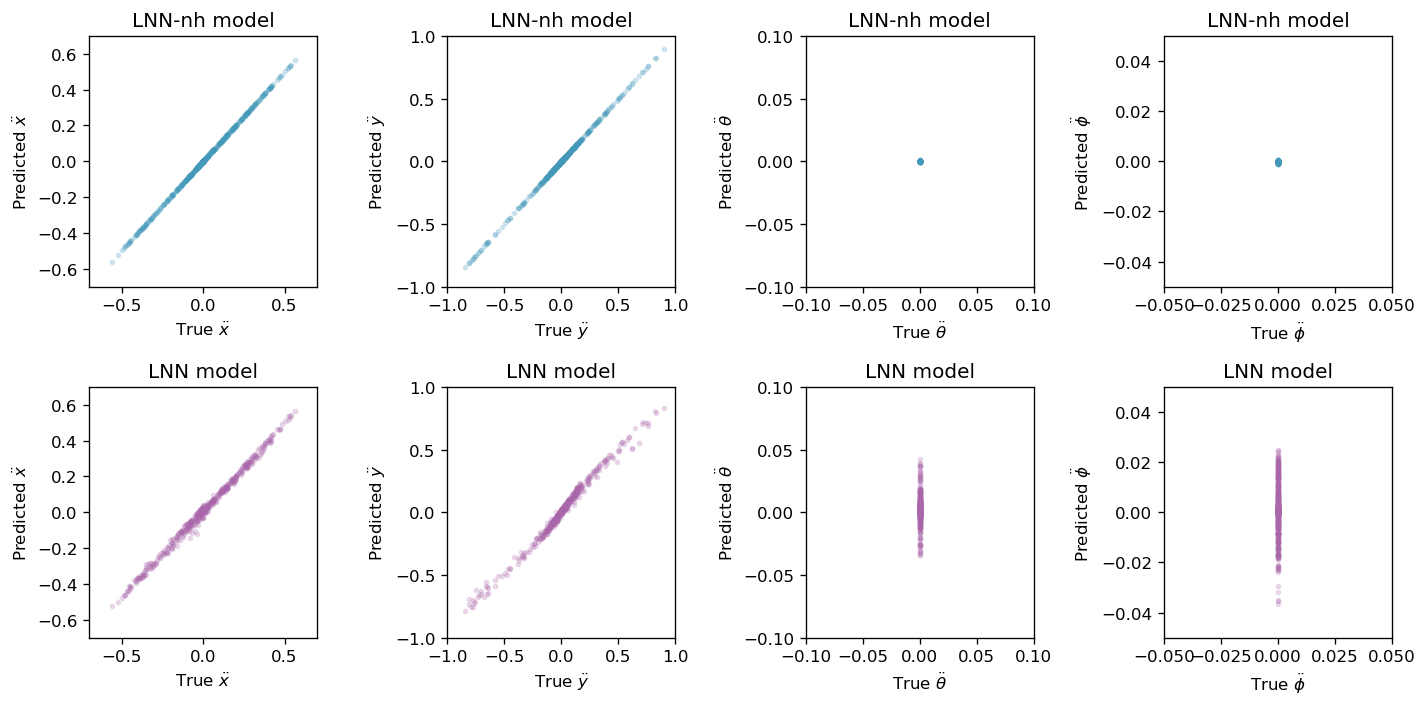}
    \caption{\label{fig:dispersiondsico} The scatter plots show five hundred true vs. corresponding learned accelerations for LNN-nh and LNN models in the wheel example.}
\end{figure}

\begin{figure}[h!]
   \begin{subfigure}[b]{0.55\linewidth}
   \centering 
   \includegraphics[width=\linewidth]{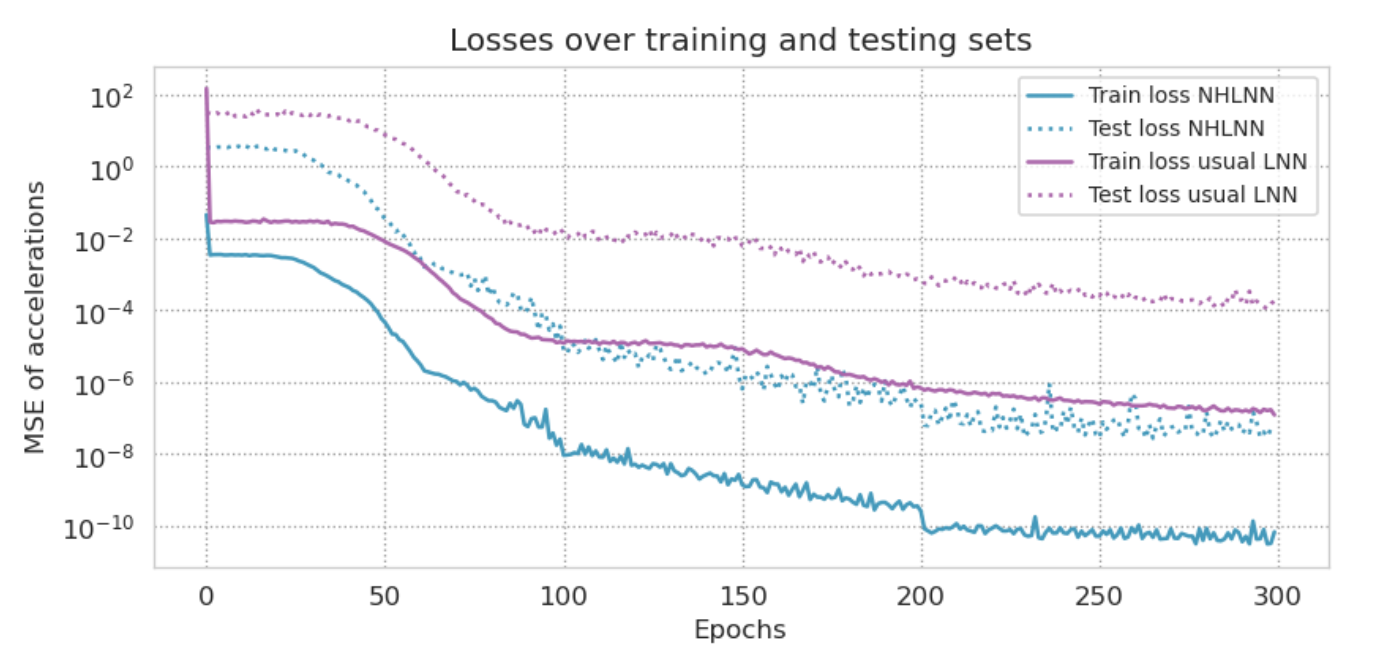} 
   \caption{Loss function}
     \label{fig:perdidadisco}
   \end{subfigure}
   \begin{subfigure}[b]{0.43\linewidth}
\includegraphics[width=\linewidth]{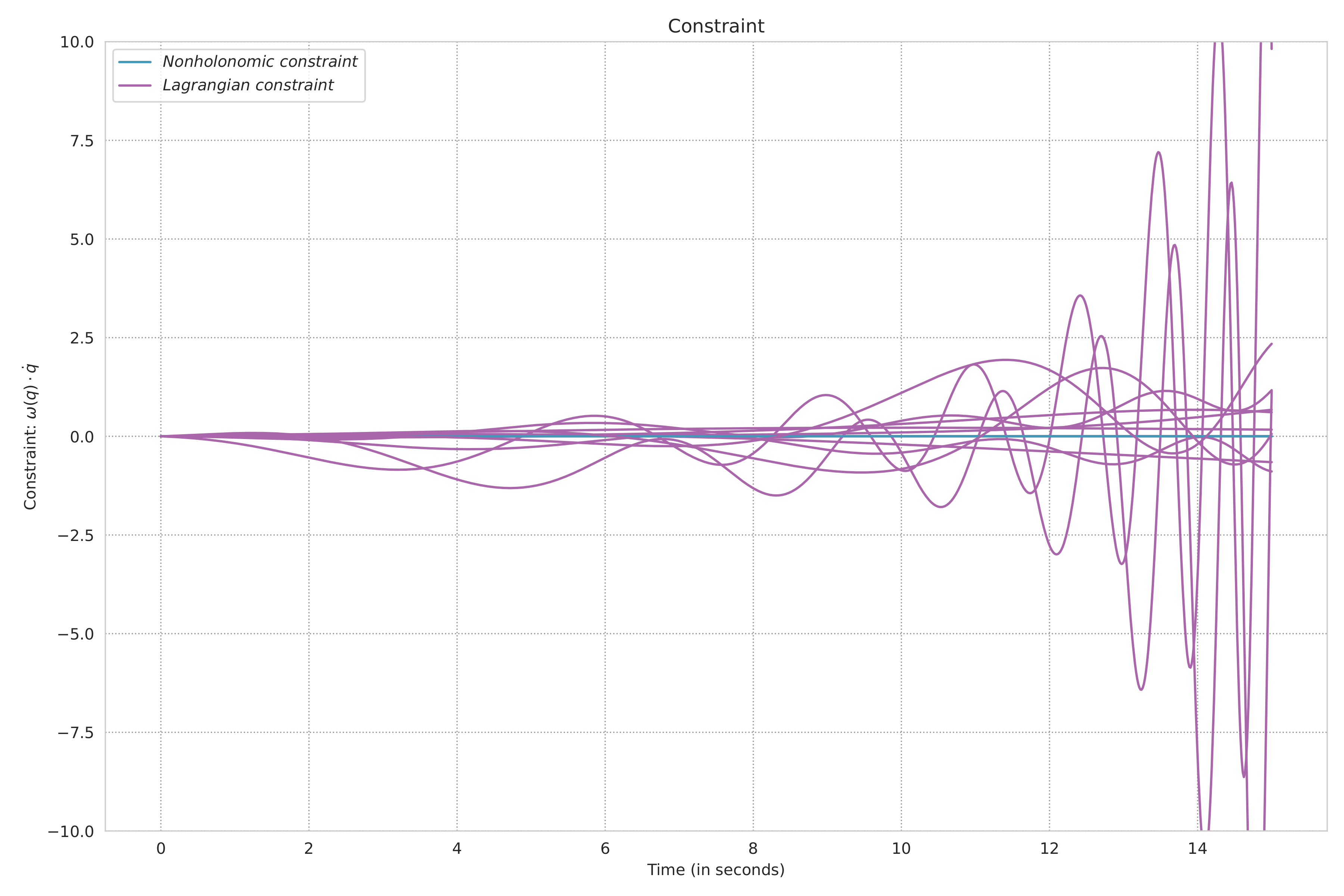}     
\caption{Nonholonomic constraint}
\label{fig:restricciondisco}
\end{subfigure}
\centering
\begin{subfigure}[b]{0.43\linewidth}
\includegraphics[width=\linewidth]{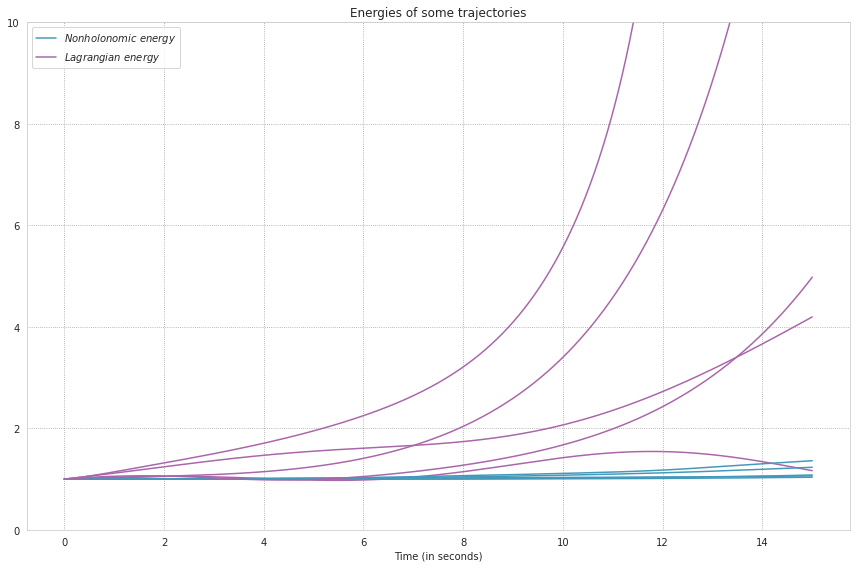}  
\caption{Total energy}
\label{fig:energiadisco}
\end{subfigure}
\begin{subfigure}[b]{0.43\linewidth}
\includegraphics[width=\linewidth]{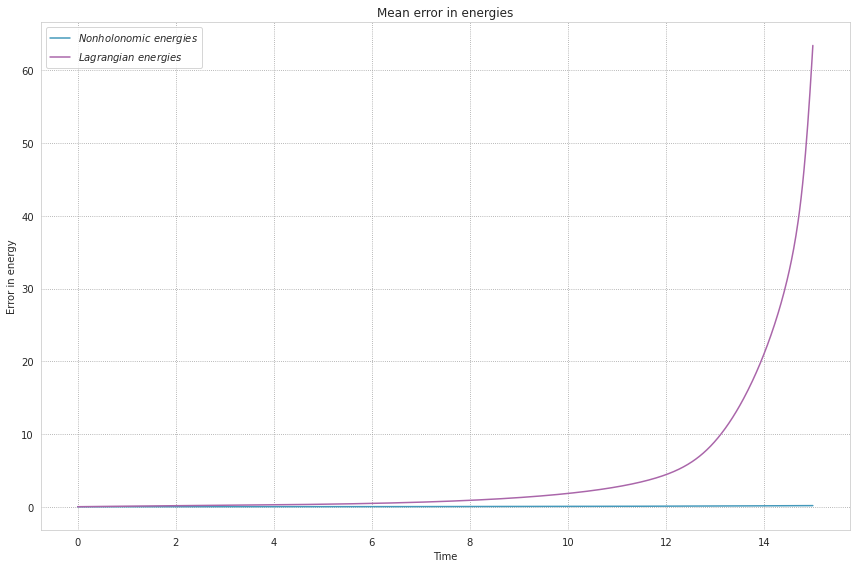}   
\caption{Mean error of energies}
\label{fig:errordisco}
\end{subfigure}

\caption{\label{fig:disco} (\ref{fig:perdidadisco}) Comparison of the loss function from the disk example for training and testing sets corresponding to LNN and LNN-nh models. Picture (\ref{fig:restricciondisco}) shows the evolution of the constraint function along five trajectories of the nonholonomic particle generated from the same initial conditions, first using the Lagrangian learned from a LNN model and second using a LNN-nh model. Picture (\ref{fig:energiadisco}) exhibits the total energy of each trajectory normalized with the corresponding constant true energy. In Picture (\ref{fig:errordisco}) can be seen the mean relative error in energy of the trajectories.}
\end{figure}

\vspace{.5cm}
Consequently, Lagrange-d'Alembert equations give rise to following system
$$\begin{cases}
\ddot{x}=-R\sin(\phi)\dot{\theta}\dot{\phi} \\
\ddot{y}=R\cos(\phi)\dot{\theta}\dot{\phi} \\
\ddot{\theta}=0 \\
\ddot{\phi}=0
\end{cases}$$
together with the constraint equations. In Figure \ref{fig:dispersiondsico} can be seen the comparison between the learned and true value of each coordinate acceleration for both models, exhibiting a major dispersion in LNN learned accelerations. 
Figure \ref{fig:disco} in turn shows the evolution of loss functions over training for testing and training sets of the LNN and LNN-nh models. The same picture also shows the performance of the learned trajectories using both systems of the energy and constraint functions.

\subsection{A system with a nonlinear nonholonomic constraint: Appell's example} 

Consider a point particle of mass $m$ moving in space subjected to a constant gravitational field of strength $g$. A configuration is then given by $(x,y,z)\in \mathbb R^3$. The Lagrangian of this system may be written as
\begin{equation*}
L(q,\dot{q})=L(x,y,z,\dot{x},\dot{y},\dot{z})=\displaystyle\frac{m}{2}(\dot{x}^2+\dot{y}^2+\dot{z}^2)-mgz,    
\end{equation*}
In addition, the particle is constrained to move according to the nonholonomic constraint 
\[
\Phi(x,y,z,\dot{x},\dot{y},\dot{z})=b^2(\dot x^2 + \dot y^2)-\dot z^2=0,
\]
for some constant $b>0$. See for instance \cite{ManuelDavidJCarlos,Appell1911} and references therein for a more detailed analysis of the system.

Unlike the linear examples, the force $f$ is now nonzero and given by $f=(0,0,-mg)$. In this case the matrix $M$ is scalar and given by
\[
M=\frac{1}{m}(4b^4\dot x^2+4b^4\dot y^2+4\dot z^2).
\]
The final equations of motion obtained by using Chetaev's principle and eliminating the Lagrange multiplier as we did in previous examples  are
\[
\begin{pmatrix}
\ddot x\\
\ddot y\\
\ddot z
\end{pmatrix}
=\frac{-b^2g}{b^4\dot x^2+b^4\dot y^2+\dot z^2}
\begin{pmatrix}
\dot x\dot z\\
\dot y\dot z\\
b^2\dot x^2+b^2\dot y^2
\end{pmatrix}.
\]

\begin{figure}[h!t]
    \centering
    \includegraphics[width=\linewidth]{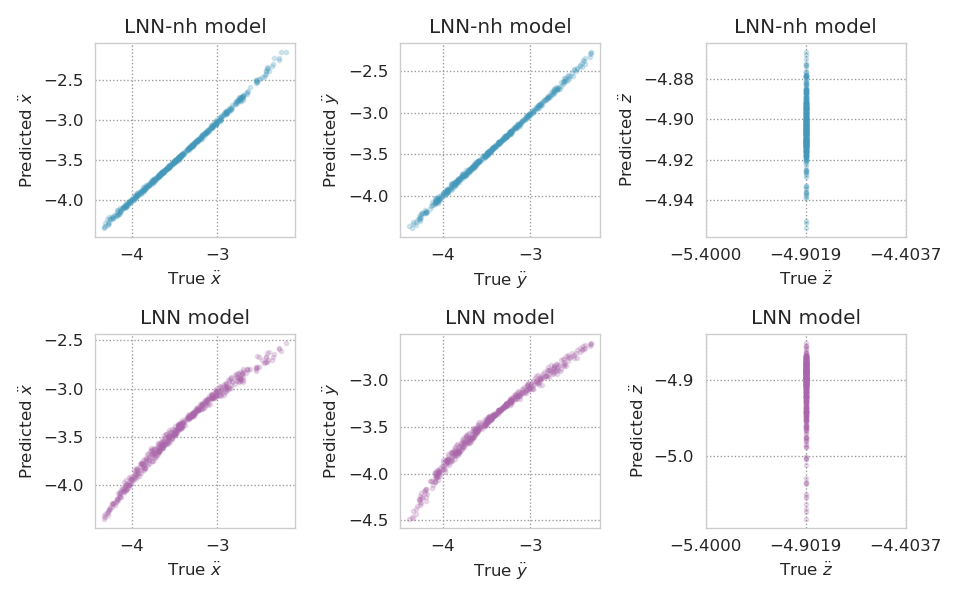}
    \caption{\label{fig:dispersionappell} The scatter plots show five hundred true vs. corresponding learned accelerations for LNN-nh and LNN models in Appell's example.}
\end{figure}

\begin{figure}[h!]
   \begin{subfigure}[b]{0.52\linewidth}
   \centering 
   \includegraphics[width=\linewidth]{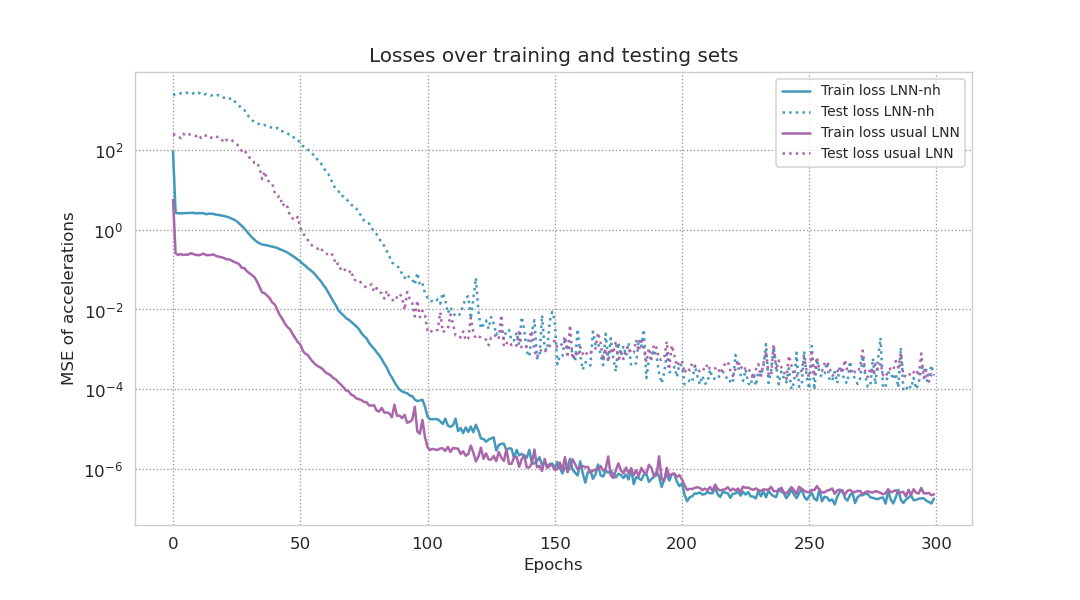} 
   \caption{Loss function}
     \label{fig:perdidaappell}
   \end{subfigure}
   \begin{subfigure}[b]{0.47\linewidth}
\includegraphics[width=\linewidth]{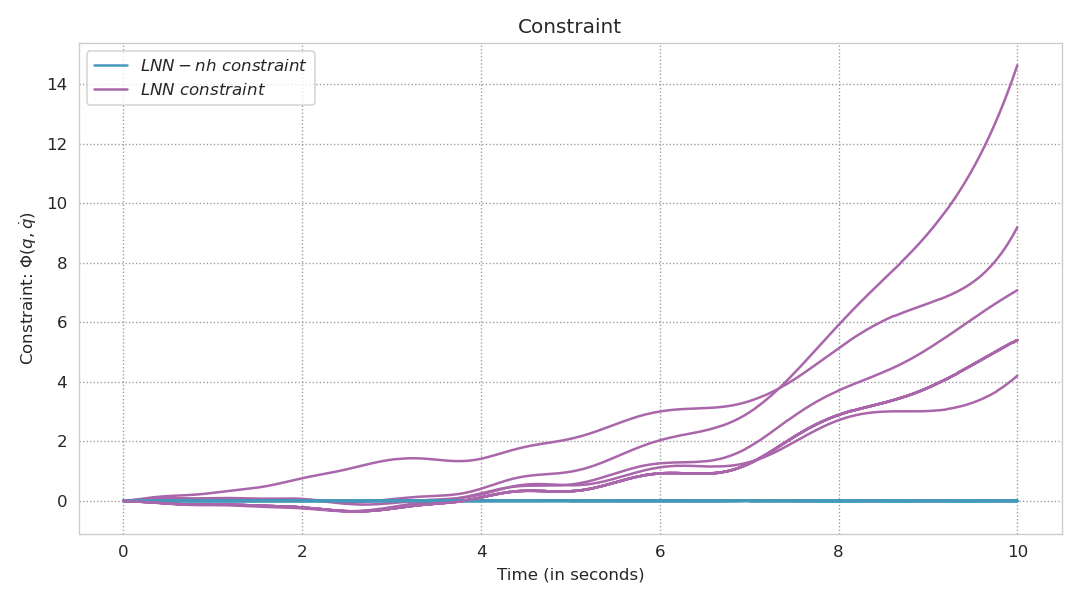}     
\caption{Nonholonomic constraint}
\label{fig:restriccionappell}
\end{subfigure}
\centering
\begin{subfigure}[b]{0.49\linewidth}
\includegraphics[width=\linewidth]{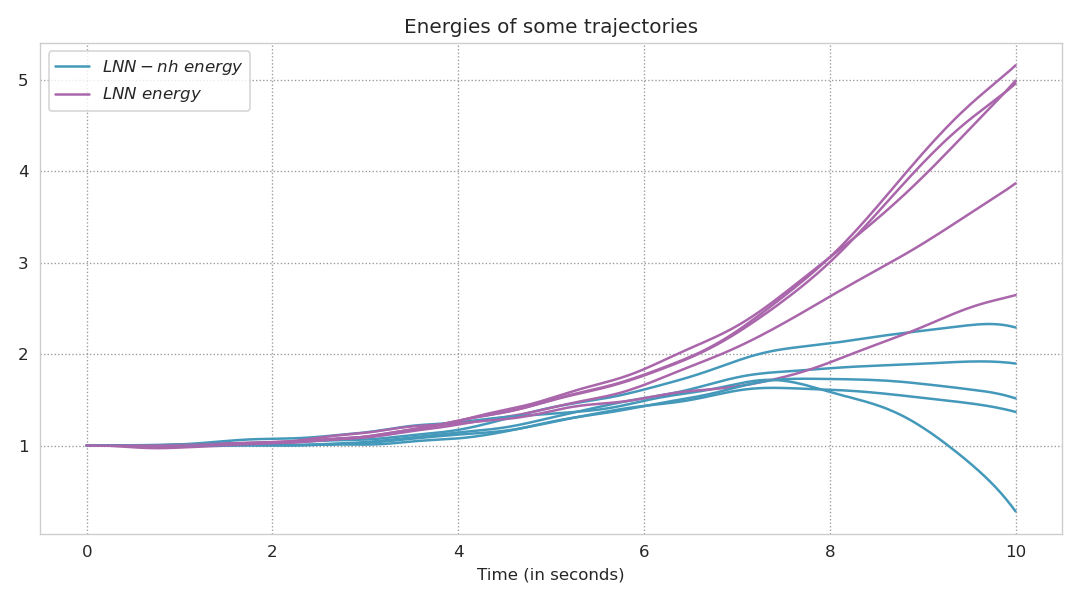}  
\caption{Total energy}
\label{fig:energiaappell}
\end{subfigure}
\begin{subfigure}[b]{0.49\linewidth}
\includegraphics[width=\linewidth]{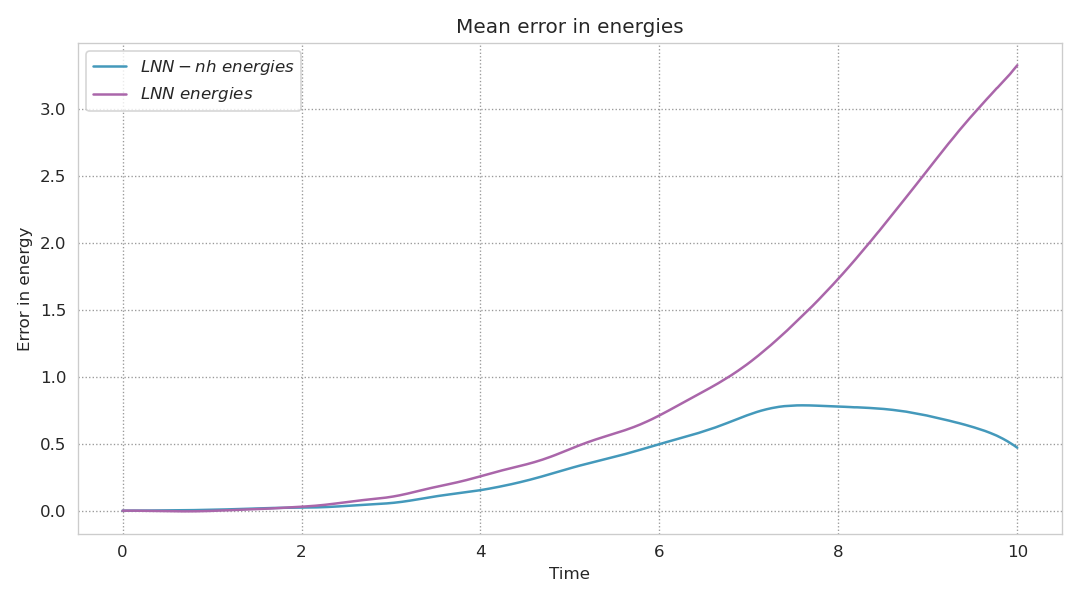}   
\caption{Mean error of energies}
\label{fig:errorappell}
\end{subfigure}

\caption{\label{fig:appell} (\ref{fig:perdidaappell}) Comparison of the loss function from Appell's example for training and testing sets corresponding to LNN and LNN-nh models. Picture (\ref{fig:restriccionappell}) shows the evolution of the constraint function along five trajectories of the nonholonomic particle generated from the same initial conditions, first using the Lagrangian learned from a LNN model and second using a LNN-nh model. Picture (\ref{fig:energiaappell}) exhibits the total energy of each trajectory normalized with the corresponding constant true energy. In Picture (\ref{fig:errorappell}) can be seen the mean relative error in energy of the trajectories.}
\end{figure}

Figure \ref{fig:dispersionappell} compares the learned and true value of each coordinate acceleration for both models. While both models exhibit reasonable predictions, we can see that accelerations predicted by the LNN model show a slight bias. 
Figure \ref{fig:appell}, in turn, shows the evolution of loss functions during training for testing and training sets of the LNN and LNN-nh models. The same figure also displays the performance of the learned trajectories using both systems of the energy and constraint functions. All computations were performed with parameters $m=1$ and $b=1$.

\section{Conclusions}

Regarding the loss graphs for training and testing for both models in the examples, we observe that although the initial values are nearly identical, the LNN-nh model shows a significantly steeper decrease in the loss function during training and testing in linear cases, and almost the same performance in the nonlinear case. By the end of training, the LNN-nh model's loss is two orders of magnitude lower than that of the LNN model in examples with linear constraints. This difference in loss is evident in the greater deviation of the LNN model's predicted accelerations from the actual values, as shown in the corresponding graphs for each example.


Concerning the conservation of energy, we can observe that the energy along LNN-nh-learned trajectories remains relatively stable over time, showing little to no increase in energy across various trajectories compared to the LNN counterpart, which exhibits fluctuations and a noticeable drift, even substantial divergence in some case, and an overall increase in energy over time. So, in general, the nonholonomic model demonstrates better stability and adherence to energy conservation principles.


In terms of comparing the constraint's behavior over time across the five trajectories of each example, the LNN model exhibits high variability, significant deviations, and sensitivity to changes over time, showing both positive and negative bias in some cases. In contrast, the LNN-nh estimation is consistently more stable, with minor fluctuations tending to remain close to zero and not showing significant differences.

\vspace{.3cm}
The examples show that our model achieves equal or lower loss in both the training and testing sets than the LNN model. It also consistently demonstrates effective energy stability and conservation, maintaining nearly constant energy levels along different trajectories. The implementations indicate that constraints from the LNN-nh model are more robust and stable over time than those from the LNN model, which are more susceptible to changes and rapidly drift away from the initial zero value of the constraint.

While the LNN model can be useful in certain contexts, it shows significant energy drift and instability in systems with nonholonomic constraints, making it less reliable for applications where energy conservation or preservation of the constraints is critical.

To summarize, across all experiments, the networks that incorporate the nonholonomic treatment of constraints into the loss function generally outperform the Lagrangian neural networks that do not consider the non-holonomic constraints. The results highlight the effectiveness of incorporating nonholonomic cons\-traints in improving Lagrangian neural network performance for systems with such kind of restrictions.


\newpage
\bibliographystyle{plain}

\bibliography{ReferRedes}

\end{document}